\pdfoutput=1
\documentclass[11pt]{article}

\usepackage{ACL2023}

\usepackage{times}
\usepackage{latexsym}

\usepackage[T1]{fontenc}
\usepackage[utf8]{inputenc}

\usepackage{microtype}

\usepackage{inconsolata}

\usepackage{graphicx}

\usepackage{makecell}

\usepackage{booktabs}
\usepackage{multirow}

\usepackage{threeparttable}
\usepackage{footmisc}
\usepackage{pifont}
\usepackage{arydshln}

\title{Pillars of Grammatical Error Correction: Comprehensive Inspection Of Contemporary Approaches In The Era of Large Language Models}

 \author{Kostiantyn Omelianchuk\thanks{\hspace{2mm}Corresponding author:\\ \href{mailto:kostiantyn.omelianchuk@grammarly.com}{kostiantyn.omelianchuk@grammarly.com}.} \\ Grammarly 
         \And
         Andrii Liubonko\\ EPAM Systems\thanks{\hspace{2mm}The work was carried out while working at Grammarly.} 
         \And
         Oleksandr Skurzhanskyi \\ Grammarly
         \AND
         Artem Chernodub \\ Grammarly
         \And
         Oleksandr Korniienko \\ Grammarly
         \And Igor Samokhin \\ Independent Researcher$^\dagger$         
         }

\begin{document}
\maketitle
\begin{abstract}

In this paper, we carry out experimental research on Grammatical Error Correction, delving into the nuances of single-model systems, comparing the efficiency of ensembling and ranking methods, and exploring the application of large language models to GEC as single-model systems, as parts of ensembles, and as ranking methods. We set new state-of-the-art performance\footnote{\url{https://nlpprogress.com/english/grammatical_error_correction.html} (Accessed 10 March 2024).\label{nlpprogress_footnote}}  with $F_{0.5}$ scores of 72.8 on CoNLL-2014-test and 81.4 on BEA-test, respectively. To support further advancements in GEC and ensure the reproducibility of our research, we make our code, trained models, and systems' outputs publicly available.\footnote{\url{https://github.com/grammarly/pillars-of-gec} 
\label{ourcode_footnote}}
\end{abstract}

\section{Introduction}

Grammatical Error Correction (GEC) is the task of correcting human text for spelling and grammatical errors. There is a wide variety of GEC approaches and model architectures. In recent years, most systems have used Transformer-based architectures \cite {Bryant_2023}. A current trend involves writing prompts for Large Language Models (LLMs) such as GPT-4 \cite{OpenAI_GPT4_2023} that would generate grammatical corrections \cite{loem-etal-2023-exploring}, \cite{coyne2023analyzing}, \cite{wu2023chatgpt}, \cite{fang2023chatgpt}.

The varied approaches within GEC each possess unique strengths and limitations. 
Combining several single-model GEC systems through ensembling or ranking may smooth out their weaknesses and lead to better overall performance \cite{susanto-etal-2014-system}. Even quite simple ensembling methods, such as majority voting \cite{tarnavskyi-etal-2022-ensembling} or logistic regression \cite{qorib-etal-2022-frustratingly}, may work surprisingly well. Combining single-model systems is also often straightforward from an implementation perspective. Because only the outputs of the models are required for many ensembling algorithms, there is no need to retrain models or perform inference passes iteratively. A further review of related work is presented in the end and near the descriptions of considered methods.

Our contributions are the following:

\textbf{1. Comprehensive comparison of GEC methods.} We reproduce, evaluate, and compare the most promising existing methods in GEC, both single-model systems and ensembles. We show that usage of ensembling methods is crucial to obtain state-of-the-art performance in GEC.

\textbf{2. Establishing new state-of-the-art baselines}. We show that simple ensembling by majority vote outperforms more complex approaches and significantly boosts performance. We push the boundaries of GEC quality and achieve new state-of-the-art results on the two most common GEC evaluation datasets: $F_{0.5}=72.8$ on CoNLL-2014-test and $F_{0.5}=81.4$ on BEA-test. 

\textbf{3. Exploring the application of LLMs for GEC.} We thoroughly investigate different scenarios for leveraging large language models (LLMs) for GEC: 1) as single-model systems in a zero-shot setting, 2) as fine-tuned single-model systems, 3) as single-model systems within ensembles, and 4) as a combining algorithm for ensembles. To the best of our knowledge, we are the first to explore using GPT-4 to rank GEC edits, which contributes to a notable improvement in the Recall of ensemble systems.

\textbf{4. Commitment to open science}. In a move toward fostering transparency and encouraging further research, we open-source all our models, their outputs on evaluation datasets, and the accompanying code.\footref{ourcode_footnote} This ensures the reproducibility of our work and provides a foundation for future advancements in the field.

\section{Data for Training and Evaluation}

We use the following GEC datasets for training models (Table \ref{training-data-table-new}): 

1. \textbf{Lang-8}, an annotated dataset from the Lang-8 Corpus of Learner English \cite{tajiri-etal-2012-tense};

2. \textbf{NUCLE}, the National University of Singapore Corpus of Learner English \cite{dahlmeier-etal-2013-building}; 

3. \textbf{FCE}, the First Certificate in English dataset \cite{yannakoudakis-etal-2011-new}; 

4. \textbf{W\&I}, the Write \& Improve Corpus \cite{bryant-etal-2019-bea} (also known as BEA-Train). We also use a larger synthetic version of Lang-8 with target sentences produced by the T5 model \cite{t5_paper};

5. \textbf{cLang-8} \cite{rothe-etal-2021-simple}, and synthetic data based on two monolingual datasets;

6. \textbf{Troy-1BW} \cite{tarnavskyi-etal-2022-ensembling}, produced from the One Billion Word Benchmark \cite{chelba2014billion};

7. \textbf{Troy-Blogs} \cite{tarnavskyi-etal-2022-ensembling}, produced from the Blog Authorship Corpus \cite{Schler2006EffectsOA}.

 \begin{table}[ht]
 %\scriptsize
 \footnotesize
 \begin{tabular}{|l|c|c|c|c|c|}
 \hline

\# & \textbf{Dataset}  &  \textbf{Part} & \textbf{\# sent.} & \textbf{\# tokens} & \textbf{\% edits} \\
\hline

1 & Lang-8          & Train          & 1.04M             & 11.86M & 42\\
\hline
2 & NUCLE       & Train          & 57.0k               & 1.16M & 62 \\
  &  & Test           & 1.3k              & 30k &  90 \\ 
\hline  
3 & FCE       & Train          & 28.0k               & 455k & 62 \\
\hline
       &          & Train          & 34.3k             & 628.7k &  67  \\ 

4 & \makecell{W\&I +\\LOCNESS}          & Dev            & 4.4k              & 85k &  64 \\
         &                & Test           & 4.5k              & 62.5k &   N/A \\
\hline
        
5 & cLang-8         & Train          & 2.37M             & 28.0M & 58\\        

 \hline
6 & Troy-1BW        & Train           &  1.2M  & 30.88M & 100 \\
\hline        
7 & Troy-Blogs       & Train           & 1.2M  & 21.49M & 100 \\
\hline

\hline
\end{tabular}
\caption{Statistics of GEC datasets used in this work for training and evaluation.}\label{training-data-table-new}
\end{table} 

For evaluation, we use current standard evaluation sets for the GEC domain: the test set from the CoNLL-2014 GEC Shared Task \cite{ng-etal-2014-conll}, and the dev and test components of the W\&I + LOCNESS Corpus from the BEA-2019 GEC Shared Task (BEA-dev and BEA-test) \cite{bryant-etal-2019-bea}. For BEA-test, submissions were made through the current competition website.\footnote{\url{https://codalab.lisn.upsaclay.fr/competitions/4057}} For each dataset, we report Precision, Recall, and $F_{0.5}$ scores. To ensure an apples-to-apples comparison with previously reported GEC results, we evaluate CONLL-2014-test with M2scorer \cite{dahlmeier-ng-2012-better}, and BEA-dev with ERRANT \cite{bryant2017automatic}.

\section{Single-Model Systems}

\begin{table*}[h!]
\small
\centering
\begin{tabular}{|c|c|ccc|ccc|ccc|}
\hline
 & & 
\multicolumn{3}{c|}{\textbf{CoNLL-2014-test}} &  \multicolumn{3}{c|}{\textbf{BEA-dev}} &  \multicolumn{3}{c|}{\textbf{BEA-test}} \\

\textbf{\#} & \textbf{System} & \textbf{Precision} & \textbf{Recall} & \textbf{$\mathbf{F_{0.5}}$} & \textbf{Precision} & \textbf{Recall} & \textbf{$\mathbf{F_{0.5}}$} & \textbf{Precision} & \textbf{Recall} & \textbf{$\mathbf{F_{0.5}}$} \\ \hline
1 & Chat-LLaMa-2-7B-ZS & 42.9 &	47.3 &	43.7 &  19.1 &	34.1 &	21.0 & - & - & -\\
2 & Chat-LLaMa-2-13B-ZS & 49.1 &	56.1 &	50.4 &  30.6 &	45.0 &	32.7  & - & - & -\\
3 & GPT-3.5-ZS & 56.2 &	57.7 & 56.5 & 37.4 &	50.6 & 39.4 & - & - & - \\
4 & GPT-3.5-CoT-ZS & 56.0 &	\textbf{58.7} &	56.5 & 36.4 & \textbf{50.8} & 38.5  & - & - & - \\
5 & GPT-4-ZS & \textbf{59.0} & 55.4 & \textbf{58.2} & \textbf{42.5} &  45.0 &  \textbf{43.0} & - & - & - \\
\hline
6 & Chat-LLaMa-2-7B-FT & 75.5 &	46.8 &	67.2	&	58.3 &	46.0	& 55.3	&	72.3 &	67.4 &	71.2 \\
7 & Chat-LLaMa-2-13B-FT & 77.3 &	45.6 &	\textbf{67.9}	&	59.8 & 46.1 &	56.4	 &	74.6	& 67.8 &	73.1 \\
8 & T5-11B & 70.9 &	\textbf{56.5} &	67.5 &		60.9 &	\textbf{51.1} &	\textbf{58.6} &		73.2 &	\textbf{71.2} &	72.8 \\
9 & UL2-20B & 73.8 &	50.4 &	67.5	&	60.5 & 48.6 & 57.7 &		75.2 &	70.0 &	74.1  \\ 
10 & GECToR-2024  & 75.0 & 44.7 & 66.0 & 64.6 & 37.2 & 56.3	& 77.7 & 59.0 &73.1 \\
11 & CTC-Copy & 72.6 &	47.0 &	65.5	&	58.3&	38.0 &	52.7	&	71.7 &	59.9 &	69.0 \\
12 & EditScorer & \textbf{78.5} &	39.4	& 65.5	&	\textbf{67.3} &	36.1 &	57.4	&	\textbf{81.0} &	56.1	& \textbf{74.4} \\
\hline

\hline
\end{tabular}

\caption{All single-model systems evaluated on CoNLL-2014-test, BEA-dev, and BEA-test datasets.}
\label{table-single-systems}
\end{table*}

\subsection{Large Language Models}
We investigate the performance of open-source models from the LLaMa-2 family \cite{touvron2023llama}, as well as two proprietary models: GPT-3.5 (Chat-GPT) and GPT-4 \cite{OpenAI_GPT4_2023}. For LLaMa, we work with four models: LLaMa-2-7B, LLaMa-2-13B, Chat-LLaMa-2-7B, and Chat-LLaMa-2-7B. We use two LLaMa-2 model sizes: 7B and 13B. If the model is pre-trained for instruction following \cite{ouyang2022training}, it is denoted as "Chat-" in the model's name. 

Chat-GPT and GPT-4 are accessed through the Microsoft Azure API. We use versions \textit{gpt-3.5-turbo-0613} and \textit{gpt-4-0613}, respectively.

We explore two scenarios for performing GEC using LLMs: zero-shot prompting (denoted as "ZS") and fine-tuning (denoted as "FT").  

\subsubsection{Zero-Shot Prompting}
\label{section:zs_single_llm}

In recent studies dedicated to prompting LLMs for GEC, it was shown that LLM models tend to produce more fluent rewrites \cite{coyne2023analyzing}. At the same time, performance measured by automated metrics such as MaxMatch \cite{dahlmeier-ng-2012-better} or ERRANT has been identified as inferior. We frequently observed that these automated metrics do not always correlate well with human scores. This makes LLMs used in zero-shot prompting mode potentially attractive, especially in conjunction with other systems in an ensemble.

For the Chat-LLaMa-2 models, we use a two-tiered prompting approach that involves setting the system prompt \textit{"You are a writing assistant. Please ensure that your responses consist only of corrected texts."} to provide the context to direct the model focus toward GEC task. Then, we push the following instruction prompt to direct the model's focus toward the GEC task: 

\begin{small}
\begin{verbatim}
Fix grammatical errors for the following text.
\end{verbatim}
\end{small} 

Temperature is set to $1$. For Chat-GPT and GPT-4 models, we employ a function-calling API with the "required" parameter. This guides the LLM to more accurately identify and correct any linguistic errors within the text or replicate the input text if it was already error-free, thus ensuring consistency in the models' responses. The instruction prompt for GPT models is:

\begin{small}
\begin{verbatim}
Fix all mistakes in the text (spelling, punctuation,
grammar, etc). If there are no errors, respond with 
the original text.
\end{verbatim}
\end{small} 

Additionally, we employ a form of the chain-of-thought (CoT) prompting \cite{wei2022chain}, which involves requesting reasoning from the model before it makes corrections by means of function calling.

\subsubsection{Fine-tuning the Large Language Models}
Fine-tuning is a mainstream method for knowledge transfer. Since we have several available annotated GEC datasets, they may be used to fine-tune LLMs \cite{zhang2023multitask,kaneko2023reducing}. 

We use three datasets for fine-tuning — NUCLE, W\&I, and cLang-8 (Table \ref{training-data-table-new}) — as they are commonly used in recent GEC research \cite{zhang2023multitask, kaneko2023reducing,loem-etal-2023-exploring}. We varied the datasets and their shares to find the best combination.

We use the Transformers library\footnote{\url{https://github.com/huggingface/transformers}} to conduct 1000–1200 updates with 250 warm-up steps, a batch size of 8, and a learning rate of $1e-5$. We fine-tune only LLaMA-2 models on next token prediction task, both autocomplete and instruction-following pre-trained versions (denoted as "Chat-"). For the Chat-LLaMA-2 models, we use the following prompt:

\begin{small}
\begin{verbatim}
Rewrite the following text to make it grammatically 
correct.
[Input text]
Result:
[Output text]
\end{verbatim}
\end{small}

Additionally, we perform an ablation study on the models' size and the usefulness of the instructions (Appendix \ref{appendix_ablation}, Table \ref{table-llama2-finetuning-instructions-usage-ablation}). Not surprisingly, our results indicate that instructions work better for "Chat" versions of models.

\subsection{Sequence-to-Sequence models}

In a sequence-to-sequence approach, GEC is considered a machine translation task, where errorful sentences correspond to the source language, and error-free sentences correspond to the target language \cite{grundkiewicz2019neural, kiyono2019empirical}. In this work, we investigate two powerful Transformer-based Seq2Seq models: the open-sourced "T5-11B" \cite{rothe-etal-2021-simple}, and "UL2-20B", the instruction-tuned version of FLAN \cite{tay2022unifying}.

T5-11B is fine-tuned on W\&I + LOCNESS train data for $500$ updates with batch size $256$ and a learning rate of $1e-4$. UL2-20B is fine-tuned on W\&I + LOCNESS train data for $300$ updates with batch size $16$ and a learning rate of $5e-5$.

\begin{table*}[h]
\footnotesize
\centering
\begin{tabular}{|c|ccc|ccc|}
\hline
 & 
\multicolumn{3}{c|}{\textbf{CoNLL-2014-test}} &  \multicolumn{3}{c|}{\textbf{BEA-test}} \\

\textbf{System name} & \textbf{Precision} & \textbf{Recall} & \textbf{$\mathbf{F_{0.5}}$} & \textbf{Precision} & \textbf{Recall} & \textbf{$\mathbf{F_{0.5}}$} \\ \hline
GECToR-RoBERTa$^{(L)}$ \cite{tarnavskyi-etal-2022-ensembling}  & 70.1  & 42.7 & 62.2 & 80.6 & 52.3  & 72.7 \\
GECToR-FT-Stage-I &  \textbf{75.2}  & 44.1   & 65.9 & \textbf{78.1} & 57.7 & 72.9\\
GECToR-FT-Stage-II (GECToR-2024)  & 75.0  & \textbf{44.7}   & \textbf{66.0} & 77.7 & \textbf{59.0} & \textbf{73.1} \\
\hline
\end{tabular}

\caption{GECToR fine-tuning experiments. We compare the performance of our fine-tuned model after stage I and stage II to the initial off-the-shelf model as a baseline.}
\label{table-gector-nucle-new}
\end{table*}

\subsection{Edit-based Systems}
Edit-based GEC systems produce explicit text changes, restoring error-free language from the errorful source text. Usually, such systems are based on encoder-only architectures and are non-autoregressive; therefore, they are less resource-consuming and more attractive for productization. In this work, we consider three publicly available open-source edit-based systems for GEC: GECToR, CTC-Copy, and EditScorer.

GECToR\footnote{\url{https://github.com/MaksTarnavskyi/gector-large}} \cite{omelianchuk-etal-2020-gector1}, \cite{tarnavskyi-etal-2022-ensembling} is a family of non-autoregressive sequence tagging GEC systems. The concept revolves around training Transformer-based, encoder-only models to generate corrective edits.

CTC-Copy\footnote{\url{https://github.com/yzhangcs/ctc-copy} \label{ctccopy_footnote}}  \cite{zhang-etal-2023-non} is another non-autoregressive text editing approach. It uses Connectionist Temporal Classification (CTC) \cite{graves2006connectionist} initially developed for automatic speech recognition and introduces a novel text editing method by modeling the editing process with latent CTC alignments. This allows more flexible editing operations to be generated.

EditScorer\footnote{\url{https://github.com/AlexeySorokin/EditScorer} \label{editscorer_footnote}} \cite{sorokin-2022-improved} splits GEC into two steps: generating and scoring edits. We consider it a single-model system approach because all edits are generated by a single-model system.

We also attempt to reproduce the Seq2Edit approach \cite{stahlberg-kumar-2020-seq2edits}, \cite{kaneko2023reducing}, but fail to achieve meaningful results. Please find more details in Appendix \ref{appendix_seq2edit}.

For GECToR, we use the top-performing model, GECToR-RoBERTa$^{(L)}$ \cite{tarnavskyi-etal-2022-ensembling}. Since this model was not trained on cLang-8 data, we additionally fine-tune it on a mix of cLang-8, BEA, Troy-1BW, and Troy-Blogs data. We leverage a multi-stage fine-tuning approach from \cite{omelianchuk-etal-2020-gector1}. In stage I, a mix of cLang-8, W\&I + LOCNESS train (BEA-train), Troy-1BW, and Troy-Blogs datasets is used for fine-tuning; in stage II, the high-quality W\&I + LOCNESS train dataset is used to finish the training. During stage I, we fine-tune the model for 5 epochs, early-stopping after 3 epochs, with each epoch equal to 10000 updates and a batch size of 256.
During stage II, we further fine-tune the model for 4 epochs, with each epoch equal to 130 updates. 
The full list of hyperparameters for fine-tuning can be found in Appendix \ref{appendix_ablation}, Table \ref{table-hyperparameters-appendix}.
We refer to this new, improved GECToR model as GECToR-2024.%\footnote{We hypothesize that this paper will be accepted before the end of 2024.}

For CTC-Copy, we use the official code\footref{ctccopy_footnote} with the RoBERTa encoder to train the English GEC model.

For EditScorer, we use the open-sourced code\footref{editscorer_footnote} for GECToR-XLNet$^{(L)}$ option from \cite{tarnavskyi-etal-2022-ensembling} to sample possible edits and stagewise decoding with the RoBERTa-Large encoder to re-score them.

\subsection{Single-Model Systems Results}

The performance of single-model GEC systems is presented in Table \ref{table-single-systems}. 

We see that all zero-shot approaches considered have $F_{0.5}$ scores lower than 60 on the CoNLL-2014-test dataset, which we assume to be a lower bound on satisfactory GEC quality. They all suffer from an overcorrecting issue \cite{fang2023chatgpt}, \cite{wu2023chatgpt} that leads to poor Precision and inferior $F_{0.5}$ scores. Notably, GPT models show consistently better results compared to LLaMa. Implementing the chain-of-thought approach doesn't improve the quality.

Among the remaining approaches — LLMs with fine-tuning, sequence-to-sequence models, and edit-based systems — we do not see a clear winner. Not surprisingly, we observe that larger models (T5-11B, UL2-20B, Chat-LLaMA-2-7B-FT, Chat-LLaMA-2-13B-FT) have slightly higher Recall compared to smaller models (GECToR-2024, CTC-Copy, EditScorer). This is expressed in 1–2\% higher $F_{0.5}$ scores on CoNLLL-2014-test; however, the values on BEA-dev and BEA-test don't show the same behavior.

\begin{figure*}[htbp]
    \centering
    \includegraphics[width=\linewidth]{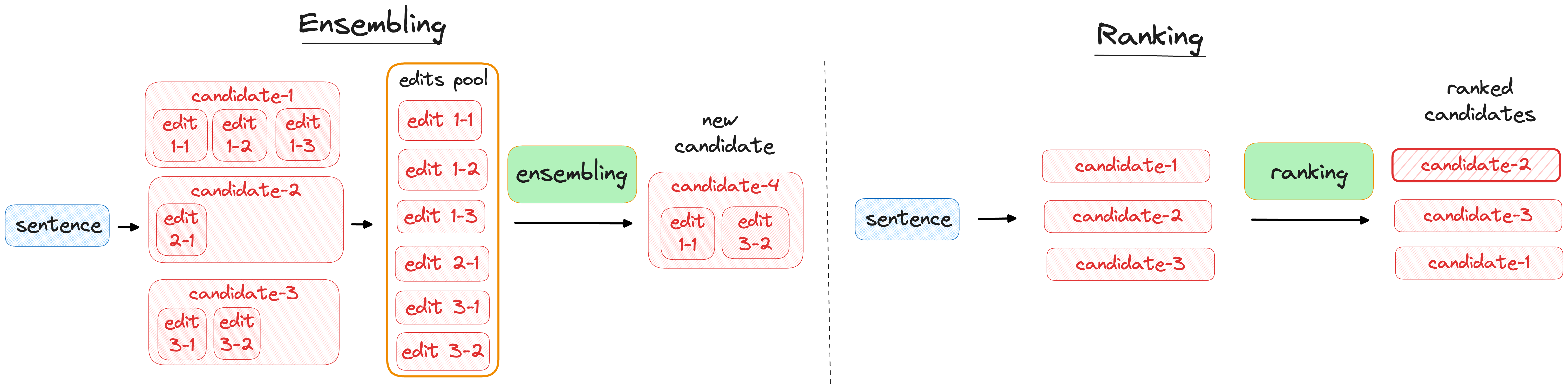}
    \caption{
        {Combining the single-model systems’ outputs. Left: In ensembling, candidates (system outputs) are aggregated on an edit level. Right: In ranking, candidates (system outputs) are aggregated on a sentence level. We consider ranking to be a special case of ensembling.}
    }
    \label{fig:ranking_vs_ensembling}
\end{figure*}

Additionally, we observe that simply scaling the model does not help achieve a breakthrough in benchmark scores. For example, a relatively small model such as GECToR-2024 ($\approx300M$ parameters) still performs well enough compared to much larger models ($\approx7-20B$ parameters). We hypothesize that the limiting factor for English GEC is the amount of high-quality data rather than model size. We have not been able to realize an $F_{0.5}$ score of more than 68\% / 59\% / 75\% on CoNLLL-2014-test / BEA-dev / BEA-test, respectively, with any single-model system approach, which is consistent with previously published results.

For GECToR, after two stages of fine-tuning, we were able to improve the $F_{0.5}$ score of the top-performing single-model model by 3.8\% on CoNLL-2014 and by 0.4\% on BEA-test, mostly due to the increase in Recall (Table \ref{table-gector-nucle-new}).

Interestingly, we see a trend where larger models exhibit diminishing returns with multi-staged training approaches. Our exploration of various training data setups reveals that a simple and straightforward approach, focusing exclusively on the W\&I + LOCNESS train dataset, performs on par with more complex configurations across both evaluation datasets. 

\section{Ensembling and Ranking of Single-Model Systems}

Combining the outputs of single-model GEC systems can improve their quality. In this paper, we explore two combining methods: ensembling and ranking (Figure \ref{fig:ranking_vs_ensembling}).

\textbf{Ensembling} combines outputs of single-model systems on an edit level. The ensemble method exploits the strengths of each model, potentially leading to more robust and accurate corrections than any single-model system could provide on its own.

\textbf{Ranking} is a special case of ensembling that combines individual outputs on a sentence level. In this approach, the performance of each system's candidate is assessed against a set of predefined criteria, and the most effective candidate is selected. Ranking maintains the internal coherence of each model's output, potentially leading to more natural and readable corrections.

\subsection{Oracle-Ensembling and Oracle-Ranking as Upper-Bound Baselines}
To set the upper-bound baseline for our experiments in combining single models, we introduce two \textit{oracle} systems: Oracle-Ensembling and Oracle-Ranking.

Oracle-Ensembling approximates an optimal combination of edits of available single-model systems. It is computationally challenging because the number of possible edit combinations grows exponentially with the number of edits. We use a heuristic to mitigate this; it optimizes Precision at the cost of reducing Recall. 

Using golden references from evaluation sets, Oracle-Ensembling works as follows:

1. Aggregate the edits from all systems into a single pool.

2. Identify and select edits that are present in both the edit pool and the available annotation. 

3. In the case of multiple annotations, we obtain a set of edits for each annotation separately. We then select the largest set of edits among the multiple annotations.

Oracle-Ranking approximates an optimal output selection for available single-model systems. Again using golden references from evaluation sets, we use M2scorer\footnote{\url{https://github.com/nusnlp/m2scorer}} to obtain $(F_{0.5},  n_{correct}, n_{proposed})$ for each system's output candidate against the available annotation. The output candidates are then sorted by $(+F_{0.5},  + n_{correct}, - n_{proposed})$ and the top one is selected. 

For our explorations into combining models' outputs, we select the seven single-model systems that show the best performance on CoNLL-2014-test (Table \ref{table-single-systems}): Chat-LLaMa-2-7B-FT, Chat-LLaMa-2-13B-FT,  T5-11B, UL2-20B, GECToR-2024, CTC-Copy, and EditScorer. As our selection criteria, we take i) systems of different types to maximize the diversity and ii) systems that have an $F_{0.5}$ score of at least 65 on CoNLL-2014-test. We refer to this set of models as "best 7".

\subsection{Ensembling by Majority Votes on Edit Spans (Unsupervised)\label{sec_majority_voting}}
%\textbf{Motivation.} 
To experiment with ensembling different GEC systems, we needed a method that is tolerant to model architecture and vocabulary size.  Ensembling by majority votes \cite{tarnavskyi-etal-2022-ensembling} on span-level edits satisfies this requirement, and it's simple to implement, so we decided to start with this approach. We use the same "best 7" set of models in our experiments.

\begin{table*}[h]
%\footnotesize
\scriptsize
\centering
\begin{tabular}{|c|ccc|ccc|ccc|}
\hline
 &  
\multicolumn{3}{c|}{\textbf{CoNLL-2014-test}} &  \multicolumn{3}{c|}{\textbf{BEA-dev}} &  \multicolumn{3}{c|}{\textbf{BEA-test}} \\

\textbf{System} &  \textbf{Precision} & \textbf{Recall} & \textbf{$\mathbf{F_{0.5}}$} & \textbf{Precision} & \textbf{Recall} & \textbf{$\mathbf{F_{0.5}}$} & \textbf{Precision} & \textbf{Recall} & \textbf{$\mathbf{F_{0.5}}$} \\ 

\hline

ESC \cite{qorib-etal-2022-frustratingly} & \textbf{81.5} & 43.8 & 69.5 & \textbf{72.9} & \textbf{40.4} & 62.8 & 86.6 & 60.9 & 79.9 \\
GRECO \cite{qorib-ng-2023-system}, var0* & 79.40 & 48.70 & 70.48 & - & - & \textbf{63.4} & 86.5 & 63.1 & 80.5 \\
GRECO \cite{qorib-ng-2023-system}, var1* & 79.60 & \textbf{49.90} & \textbf{71.12} & - & - & -  & - & - & - \\
GRECO \cite{qorib-ng-2023-system}, var2* & - & - & - & - & - & - & \textbf{86.7}  & \textbf{63.7}  & \textbf{80.8}\\

\hline

Chat-LLaMa-2-13B-FT (single-model system) & \textbf{77.3} &	45.6 &	\textbf{67.9}	&	59.8 &  46.1 &	56.4	 &	74.6	& 67.8 &	73.1 \\
UL2-20B (single-model system) & 73.8 &	\textbf{50.4} &	67.5	&	\textbf{60.5} &	\textbf{48.6} &	\textbf{57.7} &	 \textbf{75.2} &	\textbf{70.0} &	\textbf{74.1} \\

\hline

Oracle-Ensembling(best 7), baseline & \textbf{100.0} &	57.7 &	\textbf{87.2}	& \textbf{100.0} &	58.2 &	\textbf{87.4}	 &	-	& - &	- \\
Oracle-Ranking(best 7), baseline & 91.4 &	\textbf{64.2} &	84.2	&	79.6 &	\textbf{60.2} &	74.7	 &	-	& - &	- \\ 

\hline

majority-voting(best 7) &	\textbf{83.7} &	\textbf{45.7}	& \textbf{71.8}		& \textbf{71.7}	& 42.2 &	\textbf{62.9}	&	\textbf{87.3}	&64.1	& \textbf{81.4} \\ 
majority-voting(best 3) & 82.8 &	44.1 &	70.4	&	70.4 &	\textbf{43.1}	& 62.5	&	85.1 &	\textbf{64.5} &	80.0 \\ 
\hline

GRECO-ens-beam(best 7) & 77.3 & 51.6 & 70.3  & 65.5	& 47.6 & 60.9 & -  &-  &- \\
GRECO-rank(best 7) & 74.4 &	\textbf{54.2} & 69.2 & 63.2 &	\textbf{50.0} &	60.0  & -  & - & -  \\
GRECO-rank-w(best 7) & \textbf{81.6} &	49.3 &	\textbf{72.1}	&	\textbf{68.1} &	45.8 &	\textbf{62.0} & \textbf{82.0} &	\textbf{67.5} &	\textbf{78.6} \\

\hline

GPT-4-rank-prompt-a**(clust 3)**  & 72.4 &	58.3 & 69.1 & 59.7 & 52.3 & 58.1 &	-	& - &	- \\

\hline 

MAJORITY-VOTING\textsuperscript{~\ding{58}}[ majority-voting(best 7), &   & & & & & & & & \\
GRECO-rank-w(best 7) ] & 83.0 &	\textbf{48.1} &	72.5	&	70.2	& \textbf{43.9} &	62.7 &		85.6 &	\textbf{65.8} &	80.7 \\

\cdashline{1-10}[1pt/1pt]

MAJORITY-VOTING\textsuperscript{~\ding{58}}[ majority-voting(best 7), &   & & & & & & & & \\
GRECO-rank-w(best 7), GPT-4-rank-a(clust 3) ] & \textbf{83.9} &	47.5 &	\textbf{72.8} &		\textbf{70.6} &	43.5 &	\textbf{62.8}	&	\textbf{86.1} &	65.6 &	\textbf{81.1} \\

\hline
\end{tabular}
\begin{tablenotes}
\scriptsize
\item "best 7" (best 7 single-model systems): Chat-LLaMa-2-13B-FT + UL2-20B + Chat-LLaMa-2-7B-FT + EditScorer + T5-11B + CTC-Copy + GECToR-2024.
\item "best 3" (best 3 single-model systems): Chat-LLaMa-2-13B-FT + UL2-20B + Chat-LLaMa-2-7B-FT.
\item "clust 3" (clustered 3 single-model systems): Chat-LLaMa-2-13B-FT +  T5-11B +  Edit-Scorer. 
\item *In the paper \cite{qorib-ng-2023-system}, authors prepared different variants of GRECO, each of which is optimized for one test dataset.
\item **We show mean values across four GPT-4 runs with randomly shuffled single-model systems' outputs. 
\item \textsuperscript{~\ding{58}} We denote 2nd order ensembling (ensembles of ensembles) by capital letters. 
\end{tablenotes}

\caption{\label{table-ensembles} All ensembles evaluated on CoNLL-2014-test, BEA-dev, and BEA-test datasets.}
%\end{threeparttable}
\end{table*}

Our majority-vote ensembling implementation consists of the following steps:  

0. Initialization. a) Select the set of single-model systems for the ensemble. We denote the number of selected systems by $N_{sys}$. b) Set $N_{min}$, the threshold for the minimum number of edit suggestions to be accepted, $0 \leq N_{min} \leq N_{sys}$. 

1. Extract all edit suggestions from all single-model systems of the ensemble. 

2. For each edit suggestion $i$, calculate the number of single-model systems $n_i$ that triggered it.  

3. Leave only those edit suggestions that are triggered more times than the $N_{min}$ threshold: $\forall i: n_i > N_{min}$. %\geq 

4. Iteratively apply the filtered edit suggestions, beginning with the edit suggestions with the most agreement across systems (greatest $n_i$) and ending with the edit suggestions where $n_i$ is lowest. Don't apply an edit suggestion if it overlaps with one of the edits applied on a previous iteration.

\subsection{Ensembling and Ranking by GRECO Model (Supervised Quality Estimation)}

%\textbf{Motivation.} 
The quality estimation approach for combining single-model systems' outputs achieved two recent state-of-the-art results: logistic regression-based ESC (Edit-based System Combination) \cite{qorib-etal-2022-frustratingly}, and its evolution, DeBERTA-based GRECO (Grammaticality scorer for re-ranking corrections) \cite{qorib-ng-2023-system}. In this paper, we experiment with GRECO because it is open source and demonstrates state-of-the-art performance on the GEC task to the best of our knowledge\footref{nlpprogress_footnote}. GRECO was trained on the W\&I + LOCNESS training set.

We experiment with applying the publicly available GRECO model\footnote{\url{https://github.com/nusnlp/greco}} to the "best 7" set of models. We explore three ways of combining systems' outputs:

GRECO-ens-beam. We reuse beam-search implementation with beam size $k = 16$ on the edit span level.

GRECO-rank. We use GRECO to select the best single-model system's output by choosing the one with the highest score.

GRECO-rank-w. We re-weight GRECO scores for each system's output $j$ by multiplying it by a weighting coefficient $w_j$:

\begin{equation}
\forall k: w_j = \frac{n_j}{\max (n_k)},
\end{equation}

where the numerator $n_j$ is the number of systems that produce this output $j$, and the denominator  $\max (n_k)$ is the maximum number of systems for all systems' outputs. This way, we reduce the score of less frequent systems because it's not the system that is being scored/popular but rather the system's specific output (the edit).

\subsection{Ranking by GPT-4 (Zero-Shot)}

\begin{table*}[hbt!]
\small
\centering
\begin{tabular}{lcccccc}
\toprule

 & \multicolumn{3}{c}{\textbf{CoNLL-2014-test}} & \multicolumn{3}{c}{\textbf{BEA-dev}} \\
\cmidrule(lr){2-4} \cmidrule(lr){5-7}
 & \textbf{Precision} & \textbf{Recall} & $\mathbf{F_{0.5}}$ & \textbf{Precision} & \textbf{Recall} & $\mathbf{F_{0.5}}$ \\
\midrule

GPT-4-rank-prompt-a(best 7) & 70.9 ± 0.5 & 59.7 ± 0.6 & 68.4 ± 0.5 & 56.8 ± 0.3 & 53.4 ± 0.8 & 56.1 ± 0.3 \\

GPT-4-rank-prompt-b(best 7) & 69.6 ± 0.8 & 59.5 ± 0.2 & 67.3 ± 0.7 & 56.3 ± 0.5 & 53.9 ± 0.6 & 55.8 ± 0.4 \\

GPT-4-rank-prompt-a(clust 3)	& 72.4 ± 0.3 &	58.3 ± 0.6 & 69.1 ± 0.1 & 59.7 ± 0.1 & 52.3 ± 0.4 & 58.1 ± 0.1 \\

GPT-4-rank-prompt-b(clust 3) & 71.9 ± 0.4 & 58.1 ± 0.5 & 68.7 ± 0.5 & 58.7 ± 0.3 & 52.0 ± 0.5 & 57.2 ± 0.3 \\
 
\bottomrule

\end{tabular}

\begin{tablenotes}
\scriptsize
\item "best 7" (best 7 single-model systems): Chat-LLaMa-2-13B-FT + UL2-20B + Chat-LLaMa-2-7B-FT + EditScorer + T5-11B + CTC-Copy + GECToR-2024.
\item "clust 3" (clustered 3 single-model systems): Chat-LLaMa-2-13B-FT +  T5-11B +  Edit-Scorer. 
\end{tablenotes}

\caption{\label{tab:llm_ranking_3sys_7sys} LLM ranking results. We run each prompt four times with randomly shuffled outputs of single-model systems' candidates and report mean ± 2std.}

\end{table*}

%\textbf{Motivation.} 
Besides the direct application of LLMs for GEC in a zero-shot setting (we consider it in the Section \ref{section:zs_single_llm}), LLMs may be used as a combining method for ensembles. We explore GPT-4 as a ranking tool for single-model GEC systems' outputs. 

%\textbf{Experiment setup.} 
We use version \textit{gpt-4-0613} for GPT-4 with temperature $1$. We implement two prompts, "prompt-a", and "prompt-b", with slightly different goals: prompt-a aims to select the top single-model system's output among the systems' candidates, whereas prompt-b aims to perform the full ranking of the systems' candidates. They both have the same task description. For the following example of ranking three systems, it is:

\begin{small}
\begin{verbatim}
ORIGINAL:
I likes turtles very much.
EDITED:
A: I like turtles very much.
B: I likes turtles very much.
C: I like turtles very much.
\end{verbatim}
\end{small} 
But they require a different output format:\\
prompt-a (top cand.): \ \ \ \ \ \ \ \ \ \ prompt-b (ranking):

\begin{small}
\begin{verbatim}
OUTPUT:                     OUTPUT:
C                           C A B
\end{verbatim}
\end{small}

To eliminate potential positional bias, we run each prompt four times with a randomly shuffled order of single-model systems' outputs and average the performance scores.  To investigate the impact of the number of systems to be ranked, we evaluate the performance of GPT-4 on two sets of single models: "best 7" and "clust 3". 

"clust 3" refers to 3 of the 7 best single-model systems: Chat-LLaMa-2-13B-FT +  T5-11B +  Edit-Scorer. This is the subset of single-model systems from the "best 7" ensemble that provides the most distinct corrections. To select this set, we perform hierarchical clustering on TF-IDF vectors extracted from the BEA-dev dataset using a cosine similarity. The cosine similarity scores are averaged to produce a single matrix that reflects the collective performance of the single-model systems. The dendrogram illustrating the relationships between the systems based on distance is shown in Appendix \ref{appendix_ablation}, Fig. \ref{fig:systems_selections}. Based on the threshold $t=0.11$, we select the three clusters and choose Chat-LLaMa-2-13B-FT, T5-11B, Edit-Scorer to represent each. % of them. % (red dashed line on the dendrogram), 

\subsection{Ensembles of Ensembles}
%\textbf{Motivation.} 
Ensembles may themselves be combined via ensembling or ranking methods to potentially improve performance, and this is an approach we explore as well. We experiment with combining the outputs of three ensemble systems: majority-voting(best 7), GRECO-rank(best 7), and GPT-4-rank(clust 3). Here, majority-voting(best 7) was selected because it achieves the highest $F_{0.5}$ score; GRECO-rank(best 7) and GPT-4-rank(clust 3) have higher Recall and, therefore, potential to add value in an ensemble.  

The MAJORITY-VOTING algorithm (we denote second-order ensembling by capital letters) is identical to that described in \ref{sec_majority_voting}.

\subsection{Ensembles Results}

\textbf{Oracle ensembling \& ranking.}
Oracle-Ensembling shows $F_{0.5}$ scores of $87.2$/$87.4$ on CoNLL-2014-test/BEA-dev, while Oracle-Ranking performs notably worse with $F_{0.5}$ scores of $84.2$/$74.7$ and Precision of $91.4$/$79.6$ (Table \ref{table-ensembles}). This highlights the high potential for improvements on existing candidate generation and ensembling approaches, whereas ranking is more limited.

\textbf{Majority-voting ensembling.} The only hyperparameter for the method (the $N_{min}$ threshold) directly impacts the Precision/Recall balance: the higher it is set, the greater the Precision. We find that the best $N_{min}$ values for maximizing $F_{0.5}$ score are $N_{min} \approx N_{sys} / 2$. \textbf{With $\mathbf{N_{min} = 3}$, we achieve 71.8 on CoNLL-2014-test, outperforming the previous state-of-the-art result by 0.7, and 81.4 on BEA-test, setting a new state-of-the-art result.} (Table \ref{table-ensembles}, "best 7" systems ensemble).

We perform an ablation study to measure the impact of each system in the ensemble (Appendix \ref{appendix_ablation}, Table \ref{table-ablation-majority-voting}), where we remove systems one by one in the decreasing direction of $F_{0.5}$ score on the BEA-dev dataset. Our experiments show that even an ensemble combined from just the "best 3" systems (Chat-LLaMa-2-13B-FT, UL2-20B, and Chat-LLaMa-2-7B-FT) significantly improves the $F_{0.5}$ score over the UL2-20B single-model system (by 2.9\% on CoNLLL-2014-test, 4.8\% on BEA-dev, and 5.9\% on BEA-test). These results reinforce the significance of ensembling in achieving state-of-the-art performance on the GEC task. We hypothesize that majority-voting ensembling helps in mitigating the influence of noise within the data. By consolidating edits that are consistent across multiple systems (the true signal), and concurrently downplaying less prevalent and potentially inaccurate edits (the noise), the ensembling approach effectively enhances the overall quality and reliability of the output. Our experiments on BEA-dev can be found in Appendix \ref{appendix_ablation}, Table \ref{table-m7-ablation}. 

\textbf{Supervised ranking \& ensembling.}
Overall, leveraging GRECO (all variants) for combining systems' outputs leads to increased Recall at the cost of Precision. It leads to an improvement in $F_{0.5}$ score on CoNLLL-2014-test, achieving 72.1\% (+0.3\% from our best unsupervised ensemble, majority-voting(best 7)). However, results on BEA-test regressed (-2.8\% in $F_{0.5}$ score). GRECO-ens-beam did not outperform GRECO-rank-w in our experiments. 

\textbf{Zero-shot ranking.}
We observe that LLM-based ranking works better for three distinct single-model systems (clust 3) than for all seven best systems (best 7). We hypothesize that this performance disparity may be due to the increased complexity of selecting the optimal choice from a larger set of similar options. We also explain in this way the better performance of prompt-a (selection of the top candidate rewrite) than prompt-b (performing full ranking among candidate rewrites). Similar to GRECO-rank, we notice that GPT-4 favors Recall-oriented outputs, which leads to the highest Recall (58.4) on the CoNLLL-2014-test, but a suboptimal $F_{0.5}$ score. More results are presented in Table \ref{tab:llm_ranking_3sys_7sys} and in Appendix \ref{appendix_ablation}, Table \ref{tab:llm_ranking_3sys_7sys_full}.

\textbf{Ensembles of ensembles.}
Applying second-order ensembles, more specifically 
MAJORITY-VOTING[majority-voting(best 7), GRECO-rank-w(best 7), GPT-4-rank-a(clust 3)], helps to even further \textbf{push the state-of-the-art record on CoNNL-2014-test, achieving $\mathbf{F_{0.5} = 72.8}$ : $\mathbf{+1.7}$ compared to the previously highest reported result by GRECO, var1 \cite{qorib-ng-2023-system}} and $+1.0$ compared to our majority-voting(best 7) ensemble. 

\section{Related work \label{related_work}}

Large language models have demonstrated efficacy across a variety of natural language processing tasks, including GEC \cite{Bryant_2023}. The comparative analysis conducted by \cite{wu2023chatgpt} on the effectiveness of different models for GEC — ChatGPT, Grammarly, and open-sourced GECToR — reveals that ChatGPT possesses a distinctive capability to enhance textual content by not only correcting errors on a one-by-one basis but also by rephrasing original sentences, changing their structure to maintain grammatical correctness. The outcomes of human evaluations underscore the limitations of exclusively relying on automatic evaluation metrics for assessing GEC model performance, thereby positioning ChatGPT as a potentially invaluable resource for GEC applications.
 
Other research \cite{loem-etal-2023-exploring}, \cite{fang2023chatgpt} suggests that although zero-shot and few-shot chain-of-thought methodologies demonstrate promise in terms of error detection capabilities and the production of fluently corrected text, they generally underperform across the majority of error categories, thus failing to achieve high-quality outcomes in GEC.
Moreover, \cite{zhang-etal-2023-survey} delved into the customization of open-sourced foundation LLMs including LLaMA \cite{touvron2023llama} for writing assistant applications, with GEC as one of the tasks. The experimental findings indicate that instruction tuning for specific scenarios such as GEC significantly boosts the performance of LLMs and can be used to develop smaller models that outperform their larger, general-purpose counterparts.

Additionally, \cite{kaneko2023reducing} introduced a novel approach for predicting edit spans within source texts, redefining instruction-based fine-tuning as local sequence transduction tasks. This method not only reduces the length of target sequences but also diminishes the computational demands associated with inference. The study emphasizes that even high-performance LLMs such as ChatGPT struggle to generate accurate edit spans in zero-shot and few-shot scenarios, particularly in the correct generation of indexes, making this approach unstable.

Recent advancements in GEC have largely been attributed to the ensembling of outputs from individual models, as highlighted in studies by \cite{omelianchuk-etal-2020-gector1,tarnavskyi-etal-2022-ensembling}. When integrating systems with significant disparities, a system combination model is preferred over simple ensembles. This approach allows for effective integration of the strengths of various GEC systems, yielding better results than ensembles, as demonstrated in \cite{qorib-etal-2022-frustratingly}. Model outputs can be re-ranked using majority vote, as well as with the proposed GRECO model \cite{qorib-ng-2023-system}, a new state-of-the-art quality estimation model correlating more closely with the ${F_{0.5}}$ score of a corrected sentence, thus leading to a combined GEC system with a higher $F_{0.5}$ score. Additionally, this study proposes three methods for leveraging GEC quality estimation models in system combination: model-agnostic, model-agnostic with voting bias, and model-dependent methods.

\section*{Conclusions}

We don't find that any single-model system approach is dominant across all benchmarks. While in general, fine-tuning the larger models leads to higher $F_{0.5}$ scores, the 10–50x increase in model size leads to rather small improvements (up to 1–2 $F_{0.5}$ points). We hypothesize that the main bottleneck in improvement is high-quality data rather than system's architecture or model size.

To date, ensembling is crucial to overcome the limitations of single-model system approaches. Even a simple heuristic approach such as majority voting with just three single-model systems significantly boosts the quality (by 3–6 $F_{0.5}$ points). While more complex approaches (supervised ensembling or LLM zero-shot ranking) may lead to potentially better results (more specifically, show higher Recall), they usually do not lead to the target metric: $F_{0.5}$  improvement on GEC benchmarks.

Recent LLM-powered methods do not outperform other available approaches to date. However, being properly set, they can perform on par with other methods and lead to more powerful ensembles.

We've not yet reached the ceiling on the existing GEC benchmarks. Our research shows that it’s possible to improve previous records noticeably, setting the new state-of-the-art performance on two principal GEC benchmarks with $F_{0.5}$ scores of 72.8 on CoNLL-2014-test and 81.4 on BEA-test, which are improvements of +1.7 and +0.6, respectively.

In future work, we plan to explore the generation of high-quality synthetic GEC data powered by a state-of-the-art ensemble. We hypothesize that this could democratize the field by reducing the necessity of expensive training of large models to achieve a superior level of quality.

\section{Acknowledgements}
We express our gratitude to our colleagues Viacheslav Klimkov, Max Gubin, Viktor Zamaruiev for their valuable advices and support, and to Paige Schwartz for careful editing. We would like to thank the reviewers for their thoughtful comments and efforts towards improving our paper. To our communities: While we are writing this, our homeland Ukraine continues to resist the unprovoked Russian invasion. We are grateful to everyone who defends Ukraine, declares support to the people of Ukraine, and is sending aid. Thank you!

\section*{Limitations}

Firstly, our analysis was confined to the English language, potentially limiting the generalizability of our findings to other languages with potentially different error correction challenges. 

Next, our evaluation relied on two specific benchmarks using automated metrics, without incorporating human evaluation to assess the quality of the GEC. While automated metrics provide a scalable and objective means of evaluation, they may not fully capture the nuances of language that human judgment can offer.

Additionally, as we focus on ensembles, our research does not address the speed performance of the proposed systems. Therefore our findings may not provide a comprehensive view of the practicality and scalability of the proposed methods.

Lastly, the use of closed-source proprietary LLMs introduces a layer of uncertainty, as these models may undergo changes over time that are not publicly disclosed. Such changes could potentially affect the reproducibility of our results.

\bibliography{custom}
\bibliographystyle{acl_natbib}

%\clearpage
%\newpage
\appendix

\newpage
\clearpage
\section{Hierarchical clustering analysis for single-model systems \label{clustering}}

\begin{figure}[hbt!]
    \centering
    \includegraphics[width=\linewidth]{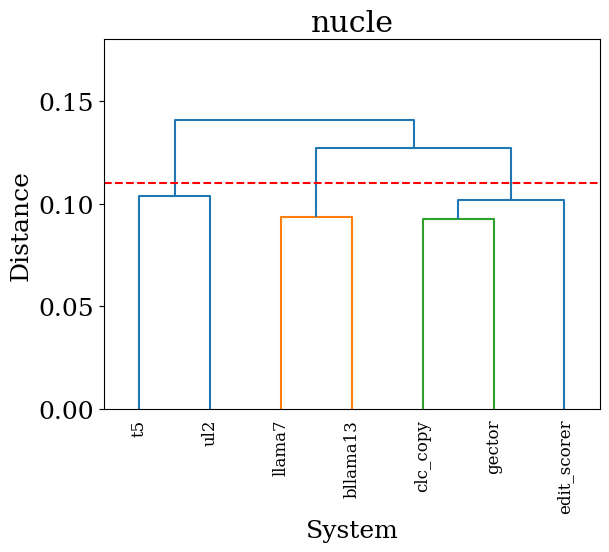}
    \caption{Dendrogram of hierarchical clustering analysis for single-model systems. The y-axis represents the distance metric used for clustering, with a red dashed line indicating the selected threshold for cluster formation ($t = 0.11$). The x-axis enumerates different systems that were analyzed. The dendrogram branches reflect the hierarchical grouping based on the proximity of distance metrics.}
    \label{fig:systems_selections}
\end{figure}

\newpage
\clearpage
\section{Unsuccessful attempt to reproduce Seq2Edit approach\label{appendix_seq2edit}}

The sequence-to-edit approach leverages the fact that in GEC, the target sentence is usually very similar to the source one. Instead of rewriting the entire sentence, it's possible to generate a list of required edits, represented as tuples: (start position, end position, replacement). \cite{stahlberg-kumar-2020-seq2edits}. We tried to re-implement the most recent approach \cite{kaneko2023reducing} that reported a high score ($F_{0.5} = 71.3\%$) on the CoNLL-2014-test. We attempted to fine-tune both T5-11B and LLaMA-2-7B models using the same set of hyperparameters that we used in our other experiments, on pairs of sentences and edits extracted from the BEA-train dataset. We were unable to get any meaningful results (our $F_{0.5}$ on CoNLL-2014-test was about 30, which is around 40 points lower than SOTA systems). Our models tended to corrupt an original sentence more often than correct it. We believe that our implementation most likely misses some crucial details required to work properly, and we encourage other researchers to reproduce and open-source the sequence-to-edit approach.  

\begin{table*}[ht]
\small
%\footnotesize
\centering
\begin{tabular}{|c|ccc|ccc|ccc|}
\hline
\textbf{Model} & \multicolumn{3}{l|}{\textbf{Datasets used for training}} & \multicolumn{3}{l|}{\textbf{CoNLL-2014-test}} & \multicolumn{3}{l|}{\textbf{BEA-dev}} \\ \hline
               & \textbf{NUCLE} & \textbf{W\&I}  & \textbf{cLang-8} & \textbf{Precision} & \textbf{Recall} & \textbf{$\mathbf{F_{0.5}}$} & \textbf{Precision} & \textbf{Recall} & \textbf{$\mathbf{F_{0.5}}$} \\ \hline
               
LLaMA-2-7B-FT      & -     & ALL & -       & 68.66   & \textbf{54.27} & 65.20  & 57.90    & \textbf{48.63} & 55.77 \\ 
LLaMA-2-7B-FT      & -     & -   & ALL & 67.25 & 50.44 & 63.05 & 57.99   & 42.11 & 53.93 \\ 
LLaMA-2-7B-FT      & ALL   & ALL & -       & 72.45   & 46.98 & 65.37 & 58.00      & 45.82 & 55.07 \\ 
Chat-LLaMa-2-7B-FT & ALL   & -   & -       & 70.39   & 36.31 & 59.42 & 50.72   & 24.51 & 41.79 \\ 
Chat-LLaMa-2-7B-FT & -     & ALL & -       & 70.45   & 52.59 & 65.97 & \textbf{59.19}   & 47.81 & \textbf{56.50}  \\ 
Chat-LLaMa-2-7B-FT & -     & ALL & 100k    & 68.94   & 52.78 & 64.96 & 57.94   & 45.53 & 54.94 \\ 
\textbf{Chat-LLaMa-2-7B-FT} & ALL   & ALL & 48k     & \textbf{75.40}    & 46.84 & \textbf{67.20}  & 58.26   & 46.03 & 55.32 \\ 
Chat-LLaMa-2-7B-FT & TP, 8k & TP, 8k & TP, 24k & 68.01   & 52.84 & 64.32 & 53.94   & 46.03 & 52.15 \\ 

\hline

\textbf{Chat-LLaMa-2-13B-FT} & ALL & ALL & 100k & \textbf{77.34} & \textbf{45.57} & \textbf{67.87} & \textbf{59.79} & \textbf{46.08} & \textbf{56.43} \\ \hline
\end{tabular}
\caption{A search of the best dataset combination for fine-tuning large language models. For fine-tuned models, different training dataset combinations were evaluated: Here, "ALL" denotes the usage of all available data for training, specific numbers (e.g., "100k") define the specific number of samples used for training, and "TP" ("true positives") denotes when only the dataset's samples containing corrections are used.}
\label{table-ablation-dataset-llm-finetuning}
\end{table*}

\begin{table*}[h]
\small
\begin{tabular}{|l|l|l|}
    \hline
    \textbf{Hyperparameter} & \textbf{Values for stage I} & \textbf{Values for stage II} \\
    \hline
    train data source & cLang8, BEA-train, 20 Troy & BEA-train  \\
    train data size & 2,897,676 & 33,618  \\
    batch\_size & 8 & 16 \\
    accumulation\_size & 32 & 16  \\
    n\_epoch & 5 & 4 \\
    patience & 3 & 3 \\
    max\_len & 50 & 50 \\
    LR & 1e-05 & 1e-05 \\
    cold\_steps\_count & 0 & 0 \\
    tp\_prob & 1 & 1 \\
    tn\_prob & 1 & 1 \\
    updates\_per\_epoch & 10000 & 0 \\
    special\_tokens\_fix & 1 & 1 \\
    transformer\_model & Roberta-large & Roberta-large \\
    Pretrained model & roberta-large\_1\_pie\_1bw\_st3 & roberta-stage1 \\
    Inference tweaks: & &\\ 
    minimum error probability & 0.65 & 0.65 \\
    Inference tweaks: & & \\ 
    confidence & 0.1 & 0.1  \\
    \hline
\end{tabular}
\caption{Hyperparameter values for the fine-tuning of GECToR-2024.}
\label{table-hyperparameters-appendix}
\end{table*}

\begin{table*}[h]
\small
%\big
%\centering
\begin{tabular}{|c|c|ccc|}
\hline
 & &   \multicolumn{3}{c|}{\textbf{BEA-dev}} \\

\textbf{System name} & $\mathbf{N_{min}}$ & \textbf{Precision} & \textbf{Recall} & c$\mathbf{F_{0.5}}$  \\ \hline

majority-voting(best 7) &	3 &	71.7 &	42.2 &	62.9 \\
\hline
\textbf{majority-voting(best 7) w/o GECToR-2024} &	3 &	73.8 &	39.1 &	\textbf{62.7} \\
majority-voting(best 7) w/o CTC-copy &	3 &	73.7 &	39.0 &	62.6 \\
majority-voting(best 7) w/o EditScorer &	3 &	72.8 &	39.5 &	62.3 \\
majority-voting(best 7) w/o T5-11B &	3	& 74.2 &	35.8	& 61.1 \\
majority-voting(best 7) w/o UL2-20B &	3	& 74.2 &	35.9 &	61.1 \\
majority-voting(best 7) w/o LlaAMA-2-7B &	3 &	74.3 &	36.2 &	61.4 \\
majority-voting(best 7) w/o LlaAMA-2-13B &	3 &	74.3 &	36.2 &	61.3 \\
\hline
majority-voting(best 6) (best 7 w/o GECToR) &	3 &	73.8 & 39.1 &	62.7 \\
\hline
majority-voting(best 6) w/o CTC-copy &	2 &	69.8 &	44.5 &	62.7 \\
majority-voting(best 6) w/o EditScorer &	2 &	69.0 &	45.3 &	62.5 \\
majority-voting(best 6) w/o T5-11B &	2 &	70.6 &	42.4 &	62.3 \\
majority-voting(best 6) w/o UL2-20B &	2 &	70.6 &	42.5 &	62.3 \\
\textbf{majority-voting(best 6) w/o Llama-2-7B}	& 2 &	71.5 &	43.2 &	\textbf{63.2} \\
majority-voting(best 6) w/o Llama-2-13B &	2 &	71.1 &	43.1 &	63.0 \\
\hline
majority-voting(best 5) (best 6 w/o Llama-2-7B) &	2 &	71.5 &	43.2 &	63.2 \\
\hline
\textbf{majority-voting(best 5) w/o CTC-copy}	&2 &	74.0&	38.8&	\textbf{62.6} \\
majority-voting(best 5) w/o EditScorer	&2&	72.6&	39.2&	62.0 \\
majority-voting(best 5) w/o T5-11B	&2&	75.1&	33.8&	60.3 \\
majority-voting(best 5) w/o UL2-20B	&2&	74.8&	34.0	&60.3 \\
majority-voting(best 5) w/o LlaMA-2-13B	&2&	74.7	&34.9&	60.8 \\
\hline
majority-voting(best 4) (best 5 w/o CTC-copy) &	2 &	74.0 &	38.8 &	62.6 \\
\hline
majority-voting(best 4) w/o EditScorer	& 1 &	66.2 &	47.9 &	61.5 \\
\textbf{majority-voting(best 4) w/o T5-11B}	& 1 &	70.4 &	43.1 &	\textbf{62.5} \\
majority-voting(best 4) w/o UL2-20B &	1 &	69.9 &	43.7 &	62.4 \\
majority-voting(best 4) w/o LlaMA-2-13B &	1 &	68.5 &	45.2 &	62.1 \\
\hline
majority-voting(best 3) (best 4 w/o T5-11B) & 1 & 70.4 & 43.1 & 62.5 \\
\hline
\textbf{majority-voting(best 3) w/o EditScorer} &	1 &	72.9 &	36.4 &	\textbf{60.7} \\
majority-voting(best 3) w/o UL2-20B &	1 &	77.0 &	28.0 &	57.0 \\
majority-voting(best 3) w/o LlaMA-2-13B	& 1 &	77.3 &	29.2 &	58.2 \\
\hline
\end{tabular}
\caption{Ablation study of removing single-model GEC systems from majority-based ensembles on BEA-dev.}
\label{table-m7-ablation}
\begin{tablenotes}
\scriptsize

\item "best 7" (best 7 single-model systems): Chat-LLaMa-2-13B-FT + UL2-20B + Chat-LLaMa-2-7B-FT + EditScorer + T5-11B + CTC-Copy + GECToR-2024.
\item "best 6" (best 6 single-model systems): Chat-LLaMa-2-13B-FT + UL2-20B + Chat-LLaMa-2-7B-FT + EditScorer + T5-11B + CTC-Copy.
\item "best 5" (best 5 single-model systems): Chat-LLaMa-2-13B-FT + UL2-20B + Chat-LLaMa-2-7B-FT + EditScorer + T5-11B.
\item "best 4" (best 4 single-model systems): Chat-LLaMa-2-13B-FT + UL2-20B + Chat-LLaMa-2-7B-FT + EditScorer.
\item "best 3" (best 3 single-model systems): Chat-LLaMa-2-13B-FT + UL2-20B + Chat-LLaMa-2-7B-FT.
\end{tablenotes}

\end{table*}

\begin{table*}[h!]
\small
\centering
\begin{tabular}{lcccccc}
%\toprule
\Xhline{4\arrayrulewidth}
 & \multicolumn{3}{c}{\textbf{CoNLL-2014-test}} & \multicolumn{3}{c}{\textbf{BEA-dev}} \\
\cmidrule(lr){2-4} \cmidrule(lr){5-7}
 & \textbf{Precision} & \textbf{Recall} & $\mathbf{F_{0.5}}$ & \textbf{Precision} & \textbf{Recall} & $\mathbf{F_{0.5}}$ \\
%\midrule
\Xhline{4\arrayrulewidth}
\multirow{1}{*}{Chat-LLaMa-2-13B-FT} 
 & 77.3 &	45.6 &	67.9	& 59.8 & 46.1 &	56.4 \\
\multirow{1}{*}{T5-11B} 
 & 70.9 &	56.5 & 67.5 & 60.9 & 51.1 & 58.6 \\
%Prompt-a (top selection) 
\Xhline{4\arrayrulewidth}
\multirow{4}{*}{GPT-4-rank-a(best 7)} 
 & 71.2 & 60.1 & 68.7 & 56.9 & 53.8 & 56.2 \\
 & 71.0 & 59.5 & 68.4 & 56.9 & 53.1 & 56.1 \\
 & 70.7 & 59.5 & 68.2 & 56.6 & 53.1 & 55.9 \\
 & 70.7 & 59.8 & 68.2 & 56.8 & 53.7 & 56.2 \\

\cdashline{1-7}[1pt/1pt]

mean ± 2std & 70.9 ± 0.5 & 59.7 ± 0.6 & 68.4 ± 0.5 & 56.8 ± 0.3 & 53.4 ± 0.8 & 56.1 ± 0.3 \\
 
\Xhline{4\arrayrulewidth}
\multirow{4}{*}{GPT-4-rank-b(best 7)} 
 & 69.2 & 59.6 & 67.0 & 56.2 & 53.8 & 55.7 \\
 & 69.6 & 59.5 & 67.3 & 56.0 & 53.5 & 55.5 \\
 & 69.5 & 59.4 & 67.2 & 56.6 & 54.0 & 56.0 \\
 & 70.2 & 59.6 & 67.8 & 56.3 & 54.2 & 55.9 \\

\cdashline{1-7}[1pt/1pt]
mean ± 2std & 69.6 ± 0.8 & 59.5 ± 0.2 & 67.3 ± 0.7 & 56.3 ± 0.5 & 53.9 ± 0.6 & 55.8 ± 0.4 \\
 
\Xhline{4\arrayrulewidth}
\multirow{4}{*}{GPT-4-rank-a(clust 3)} 
 & 72.3 & 58.4 & 69.0 & 59.8 & 52.2 & 58.1 \\
 & 72.2 & 58.6 & 69.0 & 59.7 & 52.5 & 58.1 \\
 & 72.6 & 57.9 & 69.1 & 59.7 & 52.1 & 58.0 \\
 &  72.4 & 58.4 & 69.1 & 59.7 & 52.5 & 58.1 \\

\cdashline{1-7}[1pt/1pt]
mean ± 2std	& 72.4 ± 0.3 &	58.3 ± 0.6 & 69.1 ± 0.1 & 59.7 ± 0.1 & 52.3 ± 0.4 & 58.1 ± 0.1 \\
\Xhline{4\arrayrulewidth}
\multirow{4}{*}{GPT-4-rank-b(clust 3)} 
 & 71.7 & 57.8 & 68.4 & 58.7 & 51.7 & 57.2 \\
 & 71.8 & 58.2 & 68.6 & 58.5 & 51.8 & 57.0 \\
 & 72.2 & 58.4 & 69.0 & 58.9 & 52.1 & 57.4 \\
 & 71.9 & 58.1 & 68.7 & 58.7 & 52.2 & 57.2 \\

%\hline
\cdashline{1-7}[1pt/1pt]
mean ± 2std & 71.9 ± 0.4 & 58.1 ± 0.5 & 68.7 ± 0.5 & 58.7 ± 0.3 & 52.0 ± 0.5 & 57.2 ± 0.3 \\
 
%\bottomrule

\Xhline{4\arrayrulewidth}

\end{tabular}
\caption{\label{tab:llm_ranking_3sys_7sys_full} LLM ranking for "best 7" (best 7 single-model systems): Chat-LLaMa-2-13B-FT + UL2-20B+ Chat-LLaMa-2-7B-FT + EditScorer + T5-11B + CTC-Copy + GECToR-2024) and "clust 3" (clustered 3 single-model systems: Chat-LLaMa-2-13B-FT +  T5-11B +  Edit-Scorer). We denote "prompt-a" (top candidate) as "GPT-4-rank-a", and "prompt-b" (ranking candidates) as "GPT-4-rank-b". We run each prompt four times with randomly shuffled outputs of single-model systems' candidates.}
\end{table*}

\section{Second-order ensembling of LLM-containing ensembles by aggressiveness ranking \label{appendix_aggr_rank}}
 AGGR-RANK is a ranking method that takes as input two ensembles: GPT-4-rank and an alternative ensemble. It selects GPT-4-rank under two conditions: 1) it is less "aggressive" than the alternative (it suggests fewer edited spans), and 2) it is non-trivial (edits do exist). 

The results are presented in Table \ref{apendix-table-aggr-rank}. The first system (AGGR-RANK\textsuperscript{~\ding{58}}[GPT-4-rank-a(clust 3), majority-voting(best 7)]) tends to have a higher Precision across all datasets. The second system (AGGR-RANK\textsuperscript{~\ding{58}}[GPT-4-rank-a(clust 3), GRECO-rank-w(best 7)]), despite its lower Precision, manages to achieve a slightly higher $F_{0.5}$ score on the CoNLL-2014 test dataset, suggesting that its improved Recall adequately compensates in this case. Overall, the $F_0.5$ score is generally higher for the first system on CoNLL-2014 test and BEA-test, indicating that second-order ensembling on top of the GRECO approach is the most favorable.

\begin{table*}[hbt!]
%\footnotesize
\scriptsize
\centering
\begin{tabular}{|c|ccc|ccc|ccc|}
\hline
 &  
\multicolumn{3}{c|}{\textbf{CoNLL-2014-test}} &  \multicolumn{3}{c|}{\textbf{BEA-dev}} &  \multicolumn{3}{c|}{\textbf{BEA-test}} \\

\textbf{System} &  \textbf{Precision} & \textbf{Recall} & \textbf{$\mathbf{F_{0.5}}$} & \textbf{Precision} & \textbf{Recall} & \textbf{$\mathbf{F_{0.5}}$} & \textbf{Precision} & \textbf{Recall} & \textbf{$\mathbf{F_{0.5}}$} \\ 

\hline 

AGGR-RANK\textsuperscript{~\ding{58}}[GPT-4-rank-a(clust 3),  &  & & & & & & & & \\

majority-voting(best 7)]  & \textbf{84.0} &	45.4 &	71.8	 &	\textbf{71.7} &	41.7 &	62.7 &	\textbf{87.5}	& 63.8	& \textbf{81.4} \\

\cdashline{1-10}[1pt/1pt]

AGGR-RANK\textsuperscript{~\ding{58}}[GPT-4-rank-a(clust 3),  &  & & & & & & & & \\

GRECO-rank-w(best 7)] & 81.9 &	\textbf{49.0}	& 72.2 	&	68.3 &	\textbf{45.1} &	61.9	&	82.4 &	\textbf{67.0} &	78.8 \\

\hline
\end{tabular}
\begin{tablenotes}
\scriptsize
\item "best 7" (best 7 single-model systems): Chat-LLaMa-2-13B-FT + UL2-20B + Chat-LLaMa-2-7B-FT + EditScorer + T5-11B + CTC-Copy + GECToR-2024.
\item "clust 3" (clustered 3 single-model systems): Chat-LLaMa-2-13B-FT +  T5-11B +  Edit-Scorer. 

\item *In the paper \cite{qorib-ng-2023-system}, authors prepared different variants of GRECO, each of which is optimized for one test dataset.

\item **We show mean values across four GPT-4 runs with randomly shuffled single-model systems' outputs. 
\item \textsuperscript{~\ding{58}} We denote 2nd order ensembling (ensembles of ensembles) by capital letters. 

\end{tablenotes}

\caption{\label{apendix-table-aggr-rank} Second-order ensembling by aggressiveness ranking.}
%\end{threeparttable}
\end{table*}

\section{Ablation studies\label{appendix_ablation}}

\begin{table*}[hbt!]
\scriptsize
\centering
\begin{tabular}{|c|c|ccc|ccc|}
\hline
\textbf{Model} & \textbf{Instructions} & \multicolumn{3}{c|}{\textbf{CoNLL-2014-test}} & \multicolumn{3}{c|}{\textbf{BEA-dev}} \\ 
               &  \textbf{are used}   & $\mathbf{Precision}$ & $\mathbf{Recall}$ & $\mathbf{F_{0.5}}$   & $\mathbf{Precision}$ & $\mathbf{Recall}$ & $\mathbf{F_{0.5}}$    \\ 
               
%\hline
\Xhline{4\arrayrulewidth}

LLaMA-2-7B-FT      & No                       & \textbf{69.33}   & 50.26 & 64.44 & \textbf{59.45}   & 46.29 & \textbf{56.25} \\ 
LLaMA-2-7B-FT      & Yes                      & 68.66   & \textbf{54.27} & \textbf{65.20}  & 57.9    & \textbf{48.63} & 55.77 \\ 
\hline
Chat-LLaMa-2-7B-FT & No                       & 67.53   & \textbf{53.59} & 64.19 & 58.00      & 47.37 & 55.51 \\ 
Chat-LLaMa-2-7B-FT & Yes                      & \textbf{70.45}   & 52.59 & \textbf{65.97} & \textbf{59.19}   & \textbf{47.81}& \textbf{56.50}  \\ 

%\hline
\Xhline{4\arrayrulewidth}

LLaMA-2-7B-FT & Yes & 68.66 & 54.27 & 65.20 & 57.9 & 48.63 & 55.77 \\
LLaMA-2-13B-FT & Yes & \textbf{71.49} & \textbf{55.67} & \textbf{67.65} & \textbf{60.28} & \textbf{49.26} & \textbf{57.69} \\
 \hline
Chat-LLaMa-2-7B-FT & Yes & 70.45 & 52.59 &  65.97 & \textbf{59.19} & 47.81 & 56.50 \\
Chat-LLaMa-2-13B-FT & Yes & \textbf{72.35} & \textbf{54.48} & \textbf{67.90} & 59.04 & \textbf{48.73} & \textbf{56.64} \\
%\hline
\Xhline{4\arrayrulewidth}
\end{tabular}
\caption{Ablation study on instructions' usage in fine-tuned on W\&I dataset Large Language Models.}
\label{table-llama2-finetuning-instructions-usage-ablation}
\end{table*}

\begin{table*}[hbt!]
\scriptsize
\centering
\begin{tabular}{|c|ccc|ccc|ccc|}
\hline
 &  
\multicolumn{3}{c|}{\textbf{CoNLL-2014-test}} &  \multicolumn{3}{c|}{\textbf{BEA-dev}} &  \multicolumn{3}{c|}{\textbf{BEA-test}} \\

\textbf{System} &  \textbf{Precision} & \textbf{Recall} & \textbf{$\mathbf{F_{0.5}}$} & \textbf{Precision} & \textbf{Recall} & \textbf{$\mathbf{F_{0.5}}$} & \textbf{Precision} & \textbf{Recall} & \textbf{$\mathbf{F_{0.5}}$} \\ 

\hline

majority-voting(best 7), $N_{min} = 3$ &	83.7&	45.7	& \textbf{71.8}		&71.7	&42.2&	62.9	&	87.3	&64.1	& \textbf{81.4} \\
majority-voting(best 6), $N_{min} = 3$ & 85.3	&41.7&	70.5	&	73.8	&39.1	&62.7	&	89.0 &	60.6	& 81.4 \\
majority-voting(best 5), $N_{min} = 2$ &	83.0 &	\textbf{46.3}	& 71.7	&	71.5&	\textbf{43.2}	& \textbf{63.2} &		86.4 &	\textbf{64.7} &	81.0 \\
majority-voting(best 4), $N_{min} = 2$ &	86.4 &	40.4 &	70.3	&	\textbf{74.0} &	38.8 &	62.6	&	\textbf{88.8} &	59.9 &	81.0 \\
majority-voting(best 3), $N_{min} = 1$ & 82.8 &	44.1 &	70.4	&	70.4 &	43.1	& 62.5	&	85.1 &	64.5 &	80.0 \\
majority-voting(best 2), $N_{min} = 1$ &	\textbf{86.9} &	36.3 &	67.9	 &	72.9 &	36.4 &	60.7 &		86.9 &	57.8 &	78.9 \\
\hline

\end{tabular}
\begin{tablenotes}
\scriptsize

\item "best 7" (best 7 single-model systems): Chat-LLaMa-2-13B-FT + UL2-20B + Chat-LLaMa-2-7B-FT + EditScorer + T5-11B + CTC-Copy + GECToR-2024.
\item "best 6" (best 6 single-model systems): Chat-LLaMa-2-13B-FT + UL2-20B + Chat-LLaMa-2-7B-FT + EditScorer + T5-11B + CTC-Copy.
\item "best 5" (best 5 single-model systems): Chat-LLaMa-2-13B-FT + UL2-20B + Chat-LLaMa-2-7B-FT + EditScorer + T5-11B.
\item "best 4" (best 4 single-model systems): Chat-LLaMa-2-13B-FT + UL2-20B + Chat-LLaMa-2-7B-FT + EditScorer.
\item "best 3" (best 3 single-model systems): Chat-LLaMa-2-13B-FT + UL2-20B + Chat-LLaMa-2-7B-FT.
\item "best 2" (best 2 single-model systems): Chat-LLaMa-2-13B-FT + UL2-20B.
\end{tablenotes}

\caption{\label{table-ablation-majority-voting} Ablation study for majority-voting ensembles.}
\end{table*}

\end{document}